\documentclass[onecolumn]{galbot}

\usepackage{pgfplots}
\pgfplotsset{compat=1.18}

\usepackage{wrapfig}
\usepackage{tabularx}
\usepackage{textcomp}
\usepackage{stfloats}
\usepackage{url}
\usepackage{verbatim}
\usepackage{graphicx}
\usepackage{titlesec}
\usepackage{tocloft}
\usepackage{adjustbox}
\usepackage{multirow}
\usepackage{tikz}
\usepackage{pgfplots}
\pgfplotsset{compat=1.18}
\usetikzlibrary{arrows.meta,positioning,calc,fit,backgrounds}
\usepackage{comment}
\usepackage{amsmath,amssymb}
\usepackage{colortbl}
\usepackage{color}
\usepackage{booktabs}
\usepackage{subcaption}
\usepackage{makecell}
\usepackage{array}
\usepackage{siunitx}
\usepackage{xspace}
\usepackage{float}
\usepackage{placeins}

\definecolor{slowteal}{HTML}{00706B}
\definecolor{fastamber}{HTML}{D97904}
\definecolor{softgray}{HTML}{F4F5F6}
\definecolor{linegray}{HTML}{6B7280}
\newcommand{\method}{DualWAM\xspace}
\newcommand{\sysone}{System~1\xspace}
\newcommand{\systwo}{System~2\xspace}

\title{\method: Dual-System World Action Models for\\Asynchronous Global Planning and Local Refinement}

\author[1,2,*]{Yixin Zheng}
\author[2,3,*]{Jiangran Lyu}
\author[2,4,*]{Yuntian Deng}
\author[1,2]{Kai Liu}
\authornewline[5]{Yizhou Zhou}
\author[3]{Yizhou Wang}
\author[1]{Xiaoguang Zhao}
\author[2,3,\dagger]{He Wang}
\author[2,\dagger]{Zhizheng Zhang}
\affiliation[1]{Institute of Automation, Chinese Academy of Sciences}
\affiliation[2]{Galbot}
\affiliationnewline[3]{Peking University}
\affiliation[4]{Shanghai Jiao Tong University}
\affiliation[5]{Individual Researcher}
\contribution[*]{Equal contribution}
\contribution[\dagger]{Co-corresponding authors}

\abstract{
World Action Models (WAMs) jointly generate robot actions and predict future world states, transferring priors from video pretraining to robot control. However, future visual prediction is computationally expensive, so existing WAMs often rely on long action chunks to amortize inference cost across control steps, at the cost of closed-loop responsiveness.

We present \method, a dual-system WAM that preserves broader-horizon world-action generation while enabling high-frequency closed-loop action updates by decoupling global planning and local refinement. \systwo periodically performs high-noise bidirectional denoising over a broader world-action chunk to establish a global plan, while wrist-only \sysone extracts a temporally aligned short window from the intermediate denoising state and completes low-noise refinement using the latest wrist observations, which provide action-aligned cues about local geometry, motion, and contact during interaction. The two systems operate asynchronously along a shared denoising trajectory: each global plan is reused across multiple local updates, while \sysone repeatedly incorporates fresh interaction feedback.

Across zero-shot manipulation tasks on Franka and Galbot, \method improves success over the strongest evaluated baseline by 4.5 percentage points on average, while achieving a 16.6$\times$ critical-path speedup. Further studies show that role-matched egocentric and UMI data improve success by 14 percentage points, and that the decoupled design naturally supports edge--cloud deployment with substantially lower communication overhead than the baseline.
}

\makeatletter
\patchcmd{\mymaketitle}{%
  {\color{galbotnavy!65}\contributionlist}\par
}{%
  {\color{galbotnavy!65}\contributionlist}\par
  \vskip 0.06cm
  {\small\textbf{Project Page:} \href{https://steveouo.github.io/DualWAM-Web/}{\texttt{steveouo.github.io/DualWAM-Web}}}\par
}{}{}
\apptocmd{\mymaketitle}{%
  \vspace{0pt}
    \noindent\begin{minipage}{\textwidth}
      \centering
      \captionsetup{type=figure}
      \includegraphics[width=\linewidth]{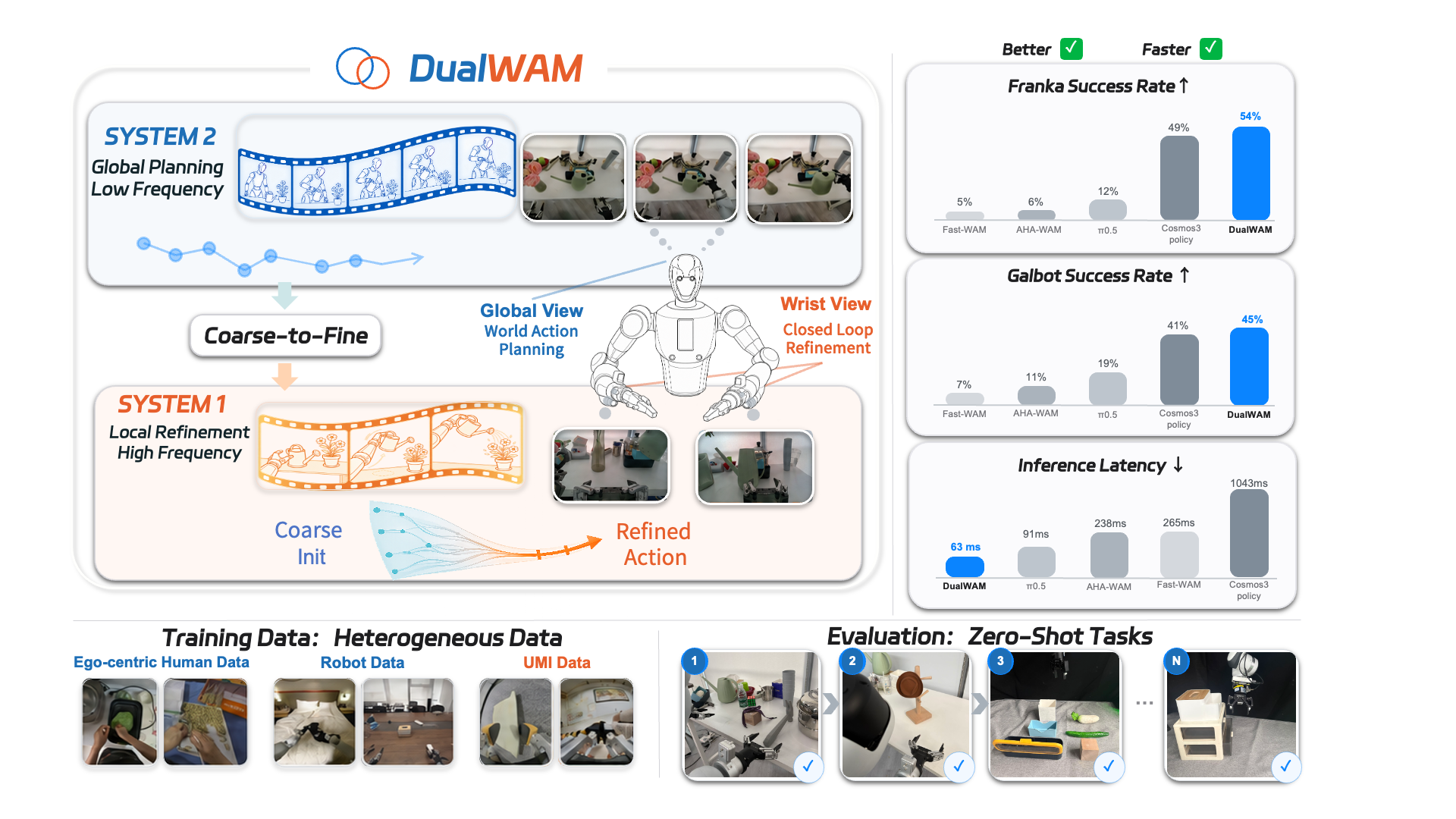}
      \vspace{1pt}
      \captionsetup{font=small}
      \captionof{figure}{\textbf{Overview.}
      \method couples low-frequency global world-action planning with high-frequency wrist-centric refinement along a shared denoising trajectory, enabling more responsive and efficient zero-shot manipulation.}
      \label{fig:teaser}
    \end{minipage}
}{}{}
\makeatother

\begin{document}
\maketitle
\pagestyle{empty}

\section{Introduction}

World Action Models (WAMs) offer a promising framework for robot manipulation by jointly predicting future world states and robot actions. By adapting pretrained video generators to control, WAMs can transfer semantic and physical priors learned from large-scale video data to robot policies~\citep{ye2026dreamzero,yuan2026fastwam,kim2026cosmospolicy}. Recent WAMs have demonstrated encouraging zero-shot generalization to tasks and interactions outside the robot training distribution~\citep{ye2026dreamzero,feng2026harmowam}.

A central challenge is that future visual prediction is computationally expensive. Existing WAMs therefore often rely on long action chunks to amortize inference cost across control steps, at the cost of closed-loop responsiveness. Executing a longer action chunk between model updates reduces how frequently new interaction feedback can affect subsequent actions. Shortening the chunk also comes with drawbacks: it requires more frequent world-action generation and encourages myopic next-chunk prediction, which can favor local appearance shortcuts over broader temporal dynamics~\citep{next-forcing}. This creates a tension between efficient global prediction and responsive closed-loop control.

Existing approaches mitigate this tension by reducing, compressing, or asynchronously decoding future predictions~\citep{yuan2026fastwam,li2026lightwam,zhang2026imagewam,lyu2026lda,cai2026ahawam,guo2026xwam}. However, they do not jointly specialize the denoising stage, visual context, and update frequency within a shared world-action generation process.

We introduce \method, a dual-system WAM that addresses this tension by decomposing world-action generation into two complementary stages: global planning and local refinement. Global planning maintains coherent structure over a broader world-action trajectory and operates at a lower frequency, while local refinement focuses on short action windows and incorporates the latest interaction feedback at every control update. Concretely, \systwo periodically applies bidirectional denoising to a broader world-action chunk, progressing from high noise to an intermediate threshold to establish the global structure of the trajectory. At each local control update, \sysone extracts a temporally aligned portion of this intermediate world-action trajectory and continues denoising it over a short local window using the latest wrist observations. The two systems operate asynchronously: each \systwo update serves multiple local control updates, while \sysone continuously incorporates fresh interaction feedback without requiring the complete trajectory to be regenerated.

This decomposition specializes the systems along three aligned dimensions. \emph{Temporally}, \systwo operates at a lower update frequency, while \sysone reacts at every local update. \emph{Generatively}, \systwo organizes a broader world-action state at high noise, whereas \sysone completes low-noise refinement over short local windows. \emph{Perceptually}, \systwo uses multi-view observations, while \sysone relies exclusively on wrist observations that naturally provide action-aligned cues about local geometry, motion, and interaction.  These role-specific observation spaces also provide a natural interface for auxiliary embodied data: egocentric videos align with global prediction, while wrist-centric Universal Manipulation Interface (UMI) demonstrations align with local refinement~\citep{chi2024universal}.

We evaluate \method on zero-shot manipulation tasks using both Franka and Galbot robots, without task-specific finetuning. \method improves success over the strongest evaluated baseline by 4.5 percentage points on average while achieving a 16.6$\times$ critical-path speedup. Further studies show that role-matched egocentric and UMI data improve success by 14 percentage points, while the decoupled design reduces average downlink traffic by a factor of 104.6 under edge--cloud execution. Our contributions are threefold. First, we introduce a dual-system WAM that assigns global planning and local refinement to complementary stages of a shared denoising trajectory. Second, we specialize these stages in update frequency, denoising regime, and visual observation space to combine coherent global generation with high-frequency closed-loop correction. Third, we validate the design through controlled ablations and zero-shot evaluation across Franka and Galbot, complemented by auxiliary-data and edge--cloud studies.

\section{Related Work}

\paragraph{World Action Models.}
World Action Models (WAMs) couple action generation with predictive modeling of future world states, either as an explicit planning interface or as an auxiliary representation for policy learning~\citep{wang2026wamsurvey,zhang2026wamtutorial}. Cascaded approaches first predict future videos, keyframes, or latent representations and subsequently derive actions from these predictions~\citep{du2023unipi,zhou2024robodreamer,liang2024dreamitate,bharadhwaj2024gen2act,zhang2026worldactionplanner,hu2024vpp,liang2025videopolicy,pai2025mimicvideo,zheng2026emergingextrinsicdexteritycluttered}. Joint WAMs instead learn future-state and action prediction within a unified autoregressive, diffusion, or flow-based generative model~\citep{wu2023gr1,cheang2024gr2,zhu2025uwm,kim2026cosmospolicy,ye2026dreamzero,li2026lingbotva}. Recent WAMs improve closed-loop control or reduce inference cost through causal generation, structured predictive states, truncated denoising, compact predictive branches, and image-, latent-, or action-centric decoding~\citep{lyu2026lda,pai2025mimicvideo,li2026lightwam,zhang2026imagewam,tian2026starry,yuan2026fastwam}. Despite these advances, most WAMs do not explicitly assign global planning and local correction to generators with different contexts, denoising regimes, observations, and update frequencies. \method introduces this specialization while retaining joint visual-action generation in both systems.

\paragraph{Asynchronous and Role-Specialized Robot Policies.}
Action chunking and diffusion-based policies provide temporally coherent multi-step control, while online chunk refinement, latent predictive representations, and asynchronous execution improve responsiveness or reduce inference overhead~\citep{zhao2023act,chi2023diffusion,black2025rtc,sendai2025a2c2,guo2026actioncontrolnet,vanjani2026damvla,bai2026latentreasoningvla}. ABot-M0.5 introduces intermediate latent actions and a dual-level architecture for unified mobility and manipulation~\citep{chen2026abot}. AHA-WAM uses a low-frequency video DiT to provide reusable context to a high-frequency action DiT~\citep{cai2026ahawam}. In contrast, both systems in \method remain WAMs along a shared denoising trajectory, allowing wrist-view dynamics to guide local action refinement. LingBot-VA 2.0 asynchronously predicts future visual latents while re-grounding subsequent predictions on the latest observation~\citep{zhang2026lingbotva2}, and X-WAM uses asynchronous noise sampling to complete action denoising earlier than multi-view RGB-D prediction~\citep{guo2026xwam}. HarmoWAM coordinates predictive and reactive action experts using world-model features, whereas DSWAM combines a vision-language subtask planner with a WAM executor~\citep{feng2026harmowam,zhu2026dswam}. \method instead decomposes world-action generation itself: \systwo establishes an overall plan, while wrist-based \sysone repeatedly refines its temporally aligned local portions using fresh observations.

\paragraph{Stage-Specialized Video Generation.}
Video diffusion models have explored combining full-sequence generation with causal or autoregressive prediction. Diffusion Forcing assigns different noise levels across tokens, Self-Forcing reduces exposure bias through model-generated contexts, and Flex-Forcing unifies bidirectional and autoregressive generation through noise-dependent attention patterns~\citep{chen2024diffusionforcing,huang2025selfforcing,ma2026flexforcing}. These methods specialize generation within a single video model. \method adapts the broader principle to robot control by assigning the high- and low-noise portions of a shared world-action trajectory to two WAMs with different temporal and perceptual roles.


\section{Method}

\begin{figure}[t]
  \centering
  \includegraphics[width=\linewidth]{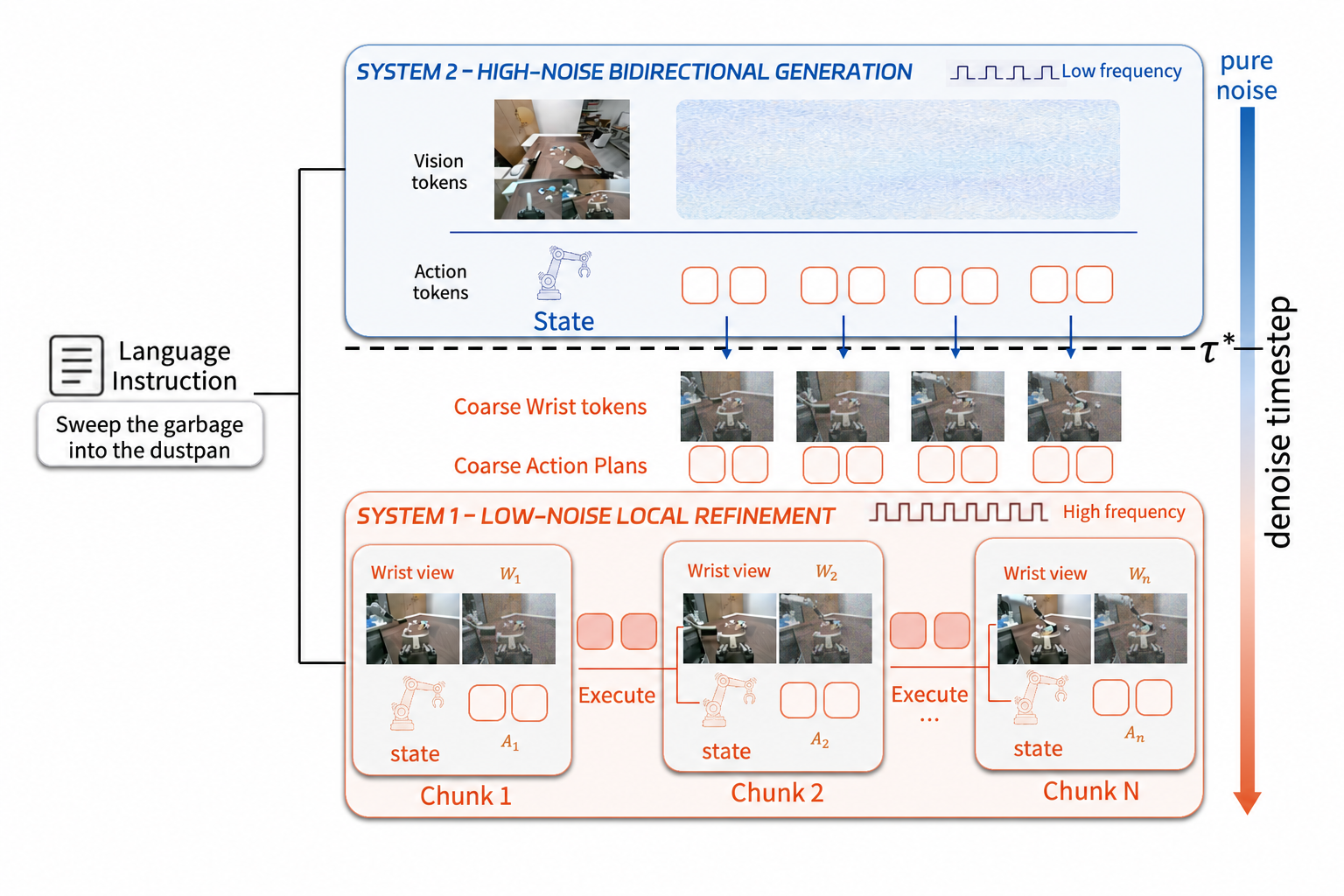}
  \caption{\textbf{\method pipeline.} \systwo denoises a complete global world-action state through the high-noise interval. At threshold $\tau$, temporally aligned wrist-action windows are passed to compact \sysone, which completes low-noise local refinement using fresh wrist observations.}
  \label{fig:pipeline}
\end{figure}

\subsection{Overview and Problem Formulation}

At control step $t$, the robot receives a language instruction $l$, proprioceptive state $q_t$, global scene observations $o_t^g$, and wrist observations $o_t^w$. Here, $o_t^g$ denotes the head-mounted or external views that capture the overall workspace, whereas $o_t^w$ denotes the wrist-mounted views used for local interaction. Given this context, \method generates a continuous stream of robot actions by assigning complementary portions of one world-action flow trajectory to two WAMs.

Let the clean global world-action target over $H_2$ video frames be
\begin{equation}
  X_0=[Z_0^g,Z_0^w,A_0],
  \qquad
  X_\sigma=(1-\sigma)X_0+\sigma\epsilon,
  \quad
  \epsilon\sim\mathcal{N}(0,I),
  \label{eq:shared_flow}
\end{equation}
where $Z_0^g$ and $Z_0^w$ are future global-view and wrist-view video latents, $A_0$ is the temporally aligned action sequence, and $\sigma\in[0,1]$ is the flow noise level, with $\sigma=1$ denoting pure noise and $\sigma=0$ the clean sample. The target rectified-flow velocity is $V=\epsilon-X_0$.

At inference, we select a handoff noise level $\tau$. \systwo denoises the complete global state from $\sigma=1$ to $\tau$, while \sysone continues from $\tau$ to $0$ over short local windows. For the $k$-th execution window, the handoff is
\begin{equation}
  U_\tau^k
  =
  \mathcal{C}_{s_k}(X_\tau;H_1)
  =
  [Z_\tau^{w,k},A_\tau^k],
  \label{eq:threshold_handoff}
\end{equation}
Here, $k$ indexes local updates, and $s_k$ is the starting video-frame offset relative to the current global trajectory, determined by execution progress. The operator $\mathcal{C}_{s_k}$ selects wrist latents covering $H_1$ video frames and actions over the corresponding time interval, accounting for temporal compression and the action sampling rate. At inference, $X_\tau$ denotes the intermediate state generated by \systwo. \sysone continues denoising its selected wrist-action variables using the shared noise coordinate.

\begin{table}[t]
  \centering
  \caption{\textbf{Complementary roles in \method.} The two WAMs specialize along aligned temporal, generative, and perceptual dimensions.}
  \label{tab:system-roles}
  \small
  \setlength{\tabcolsep}{5pt}
  \renewcommand{\arraystretch}{1.08}
  \begin{tabular}{lll}
    \toprule
    & \systwo & \sysone \\
    \midrule
    Role & Global planning & Local refinement \\
    State & $[Z^g,Z^w,A]$ & $[Z^w,A]$ \\
    Horizon (video frames) & $H_2$ & $H_1<H_2$ \\
    Noise interval & $[\tau,1]$ & $[0,\tau)$ \\
    Denoising scope & Global state & Aligned local window \\
    Update rate & Low & High \\
    Observation & Global $+$ wrist & Wrist only \\
    \bottomrule
  \end{tabular}
\end{table}

\subsection{Role-Specialized World Action Models}

\paragraph{System 2: global planning.}
\systwo processes the complete multi-view world-action state over horizon $H_2$, which provides broader temporal and visual context than a local \sysone window. At high noise levels, the visual and action trajectories remain largely undetermined. Bidirectional attention allows variables across the full prediction horizon to constrain one another, enabling \systwo to form an overall motion plan from global scene context and the available wrist observations. The resulting partially denoised trajectory $X_\tau$ is a coarse but coherent joint hypothesis over future visual states and actions; it is not independently decoded into the final executable action sequence.

\paragraph{System 1: short-horizon local refinement.}
\sysone is a compact wrist-centric WAM that refines one temporally aligned window $U_\tau^k$ at a time. At each local update, it combines this window with the latest wrist observations and robot state, then completes low-noise refinement. Because the wrist cameras are rigidly attached to the end effector, motion in the wrist-view video provides an action-aligned visual trace of executed end-effector motion. The same close-range view reveals geometry, occlusion, and contact cues that can be difficult to resolve from a wide scene view. These properties make wrist observations a natural visual input for local refinement, and \sysone therefore operates exclusively on the wrist stream. Within each window, causal temporal attention restricts every position to the current and preceding temporal positions while retaining parallel denoising across the window. As execution advances, newly acquired wrist observations trigger successive local updates, yielding high-frequency closed-loop correction. During training, we keep the first local window aligned and randomly shift the wrist-action inputs to be denoised in later windows, while keeping supervision aligned with the current execution window. The shift probability increases as execution progresses. This augmentation mimics potential misalignment between \systwo's predicted trajectory and the robot's actual execution state, making \sysone more robust during asynchronous execution.

Training samples noise levels across intervals bounded by $\tau_{\mathrm{train}}$. \sysone learns throughout the low-noise interval, allowing the inference handoff $\tau$ to be selected within that range. We train with the denoising objectives
\begin{equation}
  \mathcal{L}_i
  =
  \mathbb{E}
  \left[
    \sum_{m\in\mathcal{M}_i}
    \lambda_m^{(i)}
    \left\|
      V_m^{(i)}-\hat{V}_m^{(i)}(S_\sigma^{(i)},\sigma,c_i)
    \right\|_2^2
  \right],
  \quad
  \begin{aligned}
    \mathcal{I}_2&=[\tau_{\mathrm{train}},1], & \mathcal{M}_2&=\{g,w,a\},\\
    \mathcal{I}_1&=(0,\tau_{\mathrm{train}}], & \mathcal{M}_1&=\{w,a\},
  \end{aligned}
  \label{eq:role_objectives}
\end{equation}
Here, $g$, $w$, and $a$ denote global-view video, wrist-view video, and action variables; $c_i$ denotes each system's language, observation, and robot-state conditioning. We use $S_\sigma^{(2)}=X_\sigma$ and $V^{(2)}=\epsilon-X_0$ for \systwo. For \sysone, $S_\sigma^{(1)}$ is the temporally augmented wrist-action input, while $V^{(1)}$ denotes velocity supervision aligned with the current execution window. Temporal offsets modify the input while preserving supervision alignment. The expectation covers training examples, Gaussian noise, noise levels $\sigma\sim p_i(\sigma)$ supported on $\mathcal{I}_i$, and local-window and temporal-offset sampling for \sysone. The coefficients $\lambda_m^{(i)}$ weight the modality losses.

\subsection{Asynchronous Closed-Loop Execution}

\subsubsection{Real-time Execution of DualWAM}

World Action Models built on video generation models provide strong visual modeling and action generation capabilities. However, DualWAM is a complex architecture in which two systems operate at different execution frequencies and have explicit data dependencies. Its real-time implementation therefore raises two primary questions: \textbf{(1)} How can an appropriate asynchronous schedule reduce the model computations that must be executed serially on the action-generation critical path? \textbf{(2)} How can system-level optimizations reduce end-to-end action inference latency to the real-time regime?

\subsubsection{System-level Optimizations}

\begin{figure}[H]
  \centering
  \includegraphics[width=\linewidth]{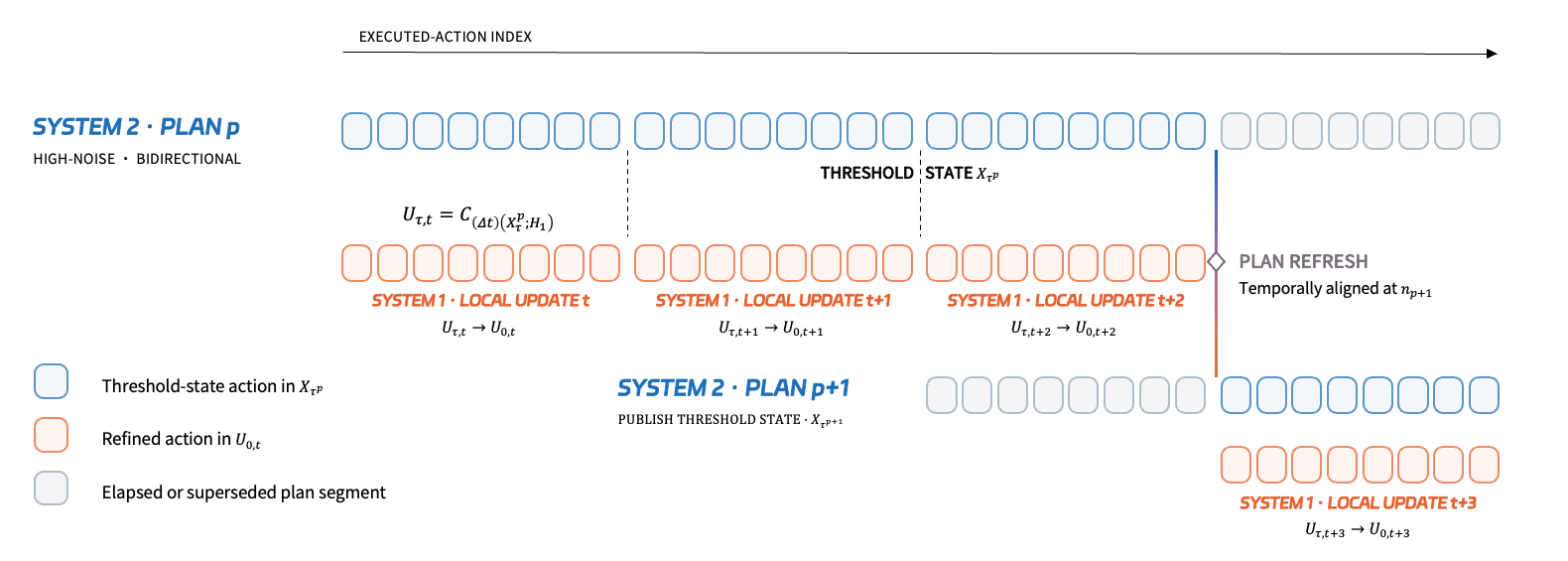}
  \caption{\textbf{Asynchronous closed-loop execution in DualWAM.} Low-frequency System 2 updates a partially denoised global world-action trajectory \(X_\tau^p\) in the background, while high-frequency System 1 repeatedly extracts the temporally aligned local window. At each plan refresh, the updated trajectory \(X_\tau^{p+1}\) is aligned with the current executed-action index. This concurrent schedule allows System 1 to maintain responsive control without waiting for System 2 to regenerate the global plan.}
  \label{fig:async}
\end{figure}

\begin{itemize}
    \item \textbf{Three-GPU CFG Parallelism} Classifier-free guidance requires two forward passes~\citep{ho2022classifier}. We separate the conditional and unconditional branches of \systwo across two GPUs in a data-parallel manner.

  \item \textbf{Asynchronous Model Execution} System 1 and System 2 execute independently and asynchronously. As shown in Figure \ref{fig:async}, System 2 updates a partially denoised global world-action trajectory in the background while System 1 refines the temporally aligned local window. This schedule prevents high-noise generation in System 2 from blocking action generation and execution in System 1. To maintain temporal alignment, System 1 skips low-noise refinement for action chunks whose execution windows have elapsed, ensuring that denoised actions remain associated with their actual execution timestamps.

\end{itemize}

\subsubsection{Implementation-level Optimizations}

\begin{itemize}
    \item \textbf{Kernel and Scheduling Optimization} Use cuDNN backend for matrix multiplication operations. We fuse the attention and integrate the activation into the MLP epilogue, reducing intermediate tensors and kernel launches. At the scheduling level, keep inputs and reusable intermediate states resident on the GPU, and connect the stages through batched and vectorized operations to reduce synchronization points and repeated H2D/D2H data transfers.

    \item \textbf{CUDA Graph Compilation.} Use CUDA Graphs to eliminate per-layer CPU dispatch and kernel-launch overhead. Capture the fixed text KV and the entire 28-layer generation stack as a single end-to-end CUDA graph, thereby further reducing runtime scheduling overhead.

    \item \textbf{Mixed-Precision Quantization} Apply W8A8 quantization to the numerically stable projection layers in the Edge decoder, while retaining BF16 for accuracy-sensitive projection layers to preserve model stability.
\end{itemize}

\subsubsection{Algorithm Improvement}

\noindent\textbf{Shared-Trajectory Denoising and Edge Timestep Sampling: $4 \rightarrow 1$.} Conventional two-stage dual-denoising pipelines typically maintain independent denoising timesteps and sampling trajectories for the two systems. Consequently, System 1 must repeat a relatively complete denoising chain on the online control path, introducing both redundant computation and a state-distribution mismatch between the two generation stages. DualWAM instead partitions a shared world-action denoising trajectory at threshold $\tau$ into complementary intervals under a common rectified-flow noise coordinate. Because the high-noise computation runs asynchronously and is amortized across multiple local control updates, the number of System 1 denoising forward passes on the online path is reduced from four to one, yielding a \textbf{\texttt{2.27x}} end-to-end speedup.

\subsubsection{Cumulative Inference Latency}

Table~\ref{tab:cumulative-inference-speedups} records measured cumulative speedups as the deployment is progressively optimized from the original single-GPU baseline to the final implementation on three NVIDIA GeForce RTX 4090 GPUs. Most of the system- and implementation-level optimizations introduce no measurable performance degradation.

\begin{table}[H]
    \centering
    \small
    \begin{tabularx}{\linewidth}{>{\raggedright\arraybackslash}X >{\raggedleft\arraybackslash}p{0.20\linewidth} >{\raggedleft\arraybackslash}p{0.20\linewidth}}
        \toprule
        \textbf{Optimization} & \textbf{Measured latency} & \textbf{Cumulative speedup} \\
        \midrule
        \textbf{Original single-GPU 4-step baseline} & \texttt{2854 ms} & \texttt{1.0x} \\
        \midrule
        \textit{System-level} & & \\
        \quad + CFG Parallelism & \texttt{1679 ms} & \texttt{1.7x} \\
        \quad + Asynchronous Execution & \texttt{501 ms} & \texttt{5.7x} \\
        \midrule
        \textit{Implementation-level} & & \\
        \quad + Kernel and Scheduling Optimization & \texttt{423 ms} & \texttt{6.8x} \\
        \quad + Torch Compile + CUDA Graphs & \texttt{280 ms} & \texttt{10.2x} \\
        \quad + Mixed-Precision Quantization & \texttt{143 ms} & \texttt{19.9x} \\
        \midrule
        \textit{Model-level} & & \\
        \quad + One-Step Sampling &
        \textbf{\texttt{63 ms}} & \textbf{\texttt{45.3x}} \\
        \bottomrule
    \end{tabularx}
    \caption{\textbf{Measured cumulative inference speedups.} Each row includes all optimizations above it.}
    \label{tab:cumulative-inference-speedups}
\end{table}

\section{Experiments}
\label{sec:experiments}

Our experiments address four questions: \textbf{(1)} How does \method compare with existing policies on zero-shot real-robot manipulation? \textbf{(2)} Does separating global planning and local refinement into two WAMs improve task performance and inference latency? \textbf{(3)} How does asynchronous execution affect closed-loop efficiency, and can the same decomposition support communication-efficient edge--cloud deployment? \textbf{(4)} Can role-matched egocentric and UMI data further improve the two systems?

\subsection{Experimental Setup}

\paragraph{Tasks and metrics.}
We evaluate zero-shot manipulation on 20 unseen real-robot tasks, with ten tasks on a Galbot G1 and ten on a Franka FR3 following the DROID setup~\citep{khazatsky2024droid}. Each policy or ablation variant is evaluated for ten trials per task, yielding 100 real-robot rollouts per embodiment. The cross-policy comparison covers both embodiments, while the architecture ablation is conducted on Franka. The edge--cloud and auxiliary-data studies are conducted on Galbot. We define a task as zero-shot when its target objects and evaluation scene do not appear in the robot training tasks and the complete primitive-skill--object--environment combination is absent from training. No task-specific demonstrations or finetuning are used. We report task progress, success rate (SR), critical-path latency, and mean task-completion time over successful rollouts. Task progress is a percentage score based on observable task-specific milestones; SR counts only rollouts that fully complete the task. Completion time measures physical execution time only and is computed as the number of executed control steps divided by the control frequency; policy inference latency is excluded. For latency evaluation, Cosmos3-Nano-Policy uses the same system and implementation optimizations as \method in Table \ref{tab:cumulative-inference-speedups}; other baselines use their official inference scripts and default sampling configurations on NVIDIA GeForce RTX 4090 GPUs. The complete prompts and progress criteria are provided in the appendix.

\begin{figure}[t]
\centering
\begin{subfigure}{0.95\linewidth}
  \centering
  \includegraphics[width=\linewidth]{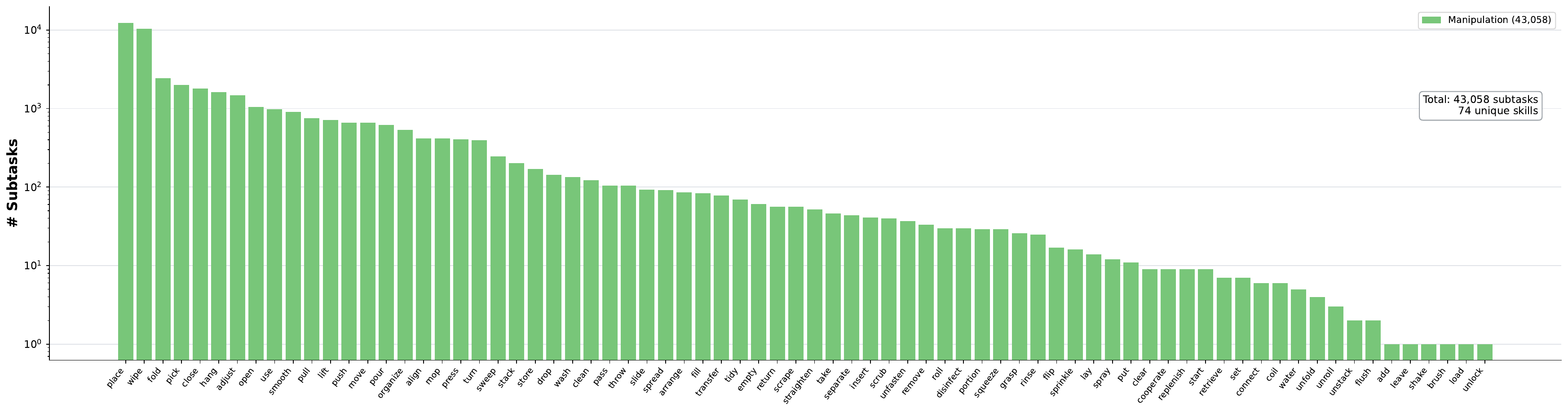}
  \phantomsubcaption\label{fig:galbot-skill-distribution}
\end{subfigure}
\vspace{2pt}
\begin{subfigure}{0.82\linewidth}
  \centering
  \includegraphics[width=\linewidth]{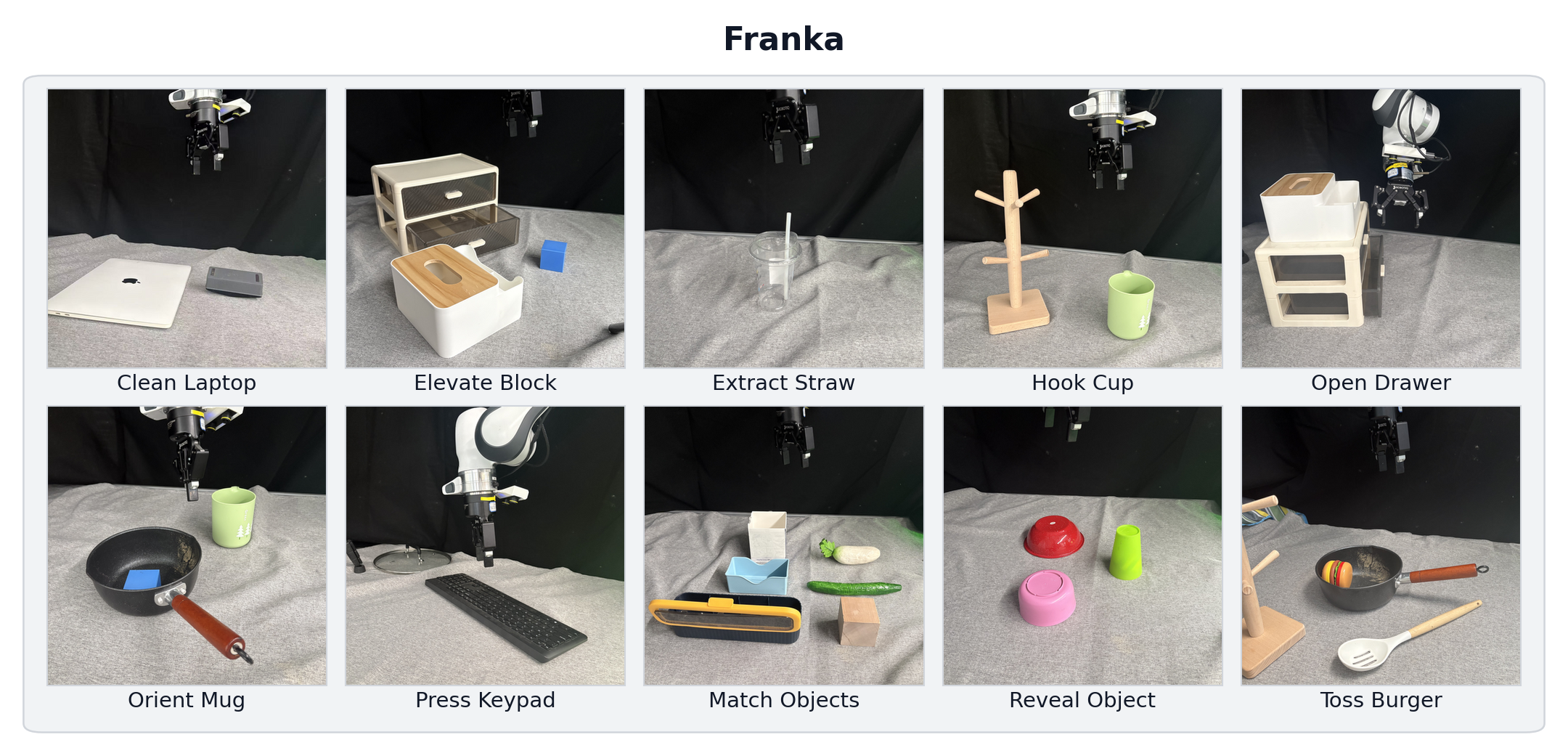}
  \phantomsubcaption\label{fig:franka-task-gallery}
\end{subfigure}
\vspace{2pt}
\begin{subfigure}{0.82\linewidth}
  \centering
  \includegraphics[width=\linewidth]{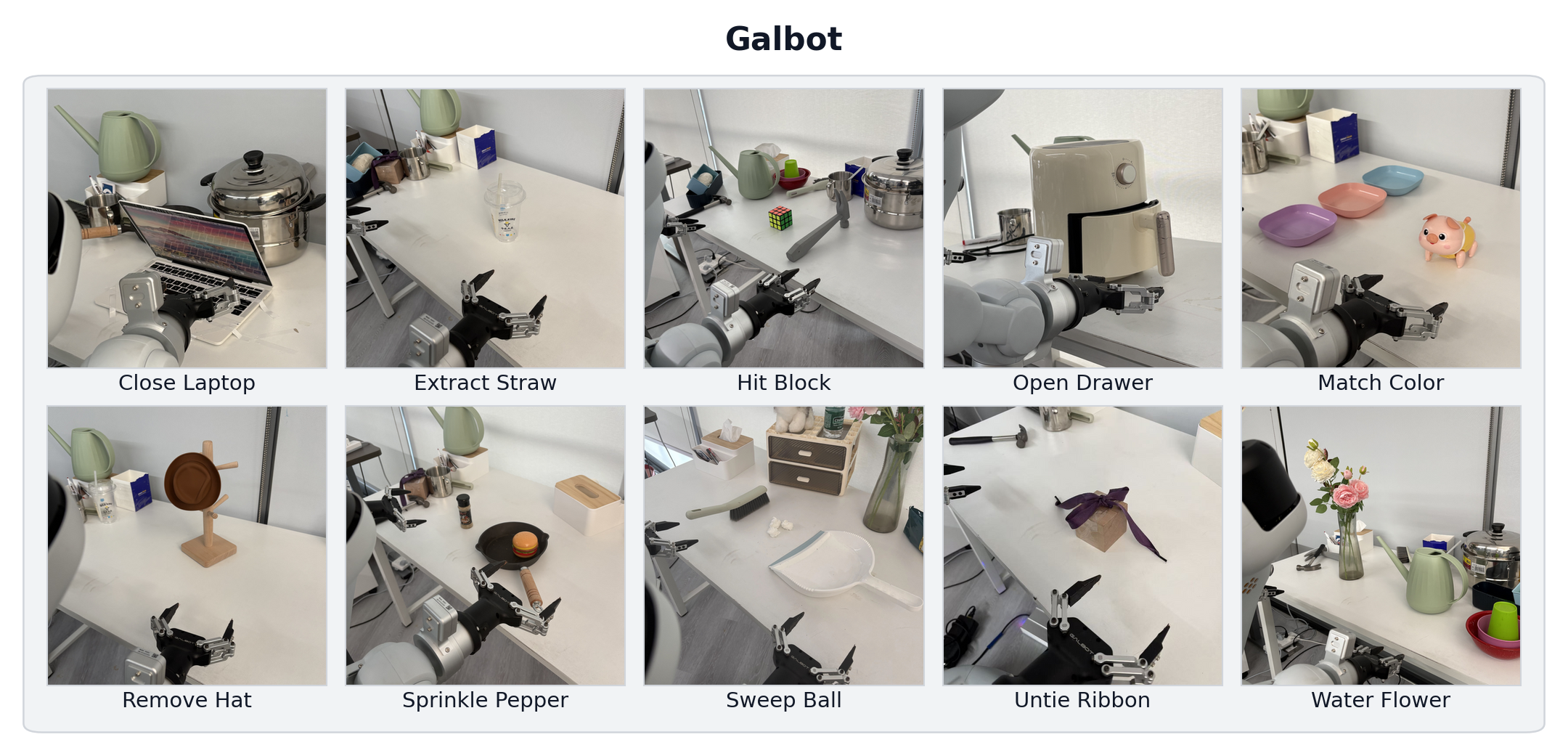}
  \phantomsubcaption\label{fig:galbot-task-gallery}
\end{subfigure}
\caption{\textbf{Training coverage and zero-shot evaluation tasks.} Top: distribution of 43,058 annotated subtasks across 74 manipulation skills in the self-collected Galbot training corpus. Middle and bottom: initial configurations for the ten Franka FR3 and ten Galbot G1 evaluation tasks, respectively.}
\label{fig:data-and-task-coverage}
\end{figure}

The self-collected Galbot robot corpus contains 43,058 annotated manipulation subtasks spanning 74 skill labels. Figure~\ref{fig:data-and-task-coverage} summarizes this training coverage together with the two zero-shot evaluation suites.

\FloatBarrier

\paragraph{Implementation details.}
For the Galbot experiments, we initialize \systwo from Cosmos3-Nano and \sysone from Cosmos3-Edge~\citep{nvidia2026cosmos3}, reinitializing the action-related parameters of both models to accommodate the Galbot action representation. For the Franka experiments, we initialize \systwo directly from Cosmos3-Nano-Policy. \sysone follows the same initialization strategy as in the Galbot setting. Both systems are trained for 10k iterations with a global batch size of 1,024 on eight nodes, each equipped with eight NVIDIA H200 GPUs. Complete implementation details and hyperparameter settings are provided in Appendix~\ref{sec:additional-implementation-details}.

\paragraph{Auxiliary data.}
The heterogeneous-data study uses two self-collected corpora in addition to robot trajectories. The egocentric corpus contains approximately 1,800 hours of human interaction video and is used to train \systwo. The wrist-centric UMI corpus contains approximately 1,200 hours of demonstrations and is used to train \sysone.

\paragraph{Baselines.}
We compare four baseline families. $\pi_{0.5}$ is an open-world VLA trained
through heterogeneous co-training~\citep{black2025pi05}.
Cosmos3-Nano-Policy post-trains the Cosmos3-Nano omnimodal world model into an
action policy~\citep{nvidia2026cosmos3}. Fast-WAM retains video co-training but
removes future prediction at inference~\citep{yuan2026fastwam}. AHA-WAM pairs
low-frequency video generation with high-frequency action prediction
asynchronously~\citep{cai2026ahawam}. For the Franka experiments, we use the
publicly released $\pi_{0.5}$ and Cosmos3-Nano-Policy checkpoints
trained on DROID, and train Fast-WAM and AHA-WAM using their official scripts and
configurations. For the Galbot experiments, we retrain every baseline on the
same data with the same batch size, initializing from each method's publicly
released bimanual checkpoint when available. We train $\pi_{0.5}$ for 50k
optimization steps and all other baselines for 10k steps.

\subsection{How Does DualWAM Compare with Existing Policies?}

\begin{figure}[t]
\centering
\includegraphics[width=\linewidth]{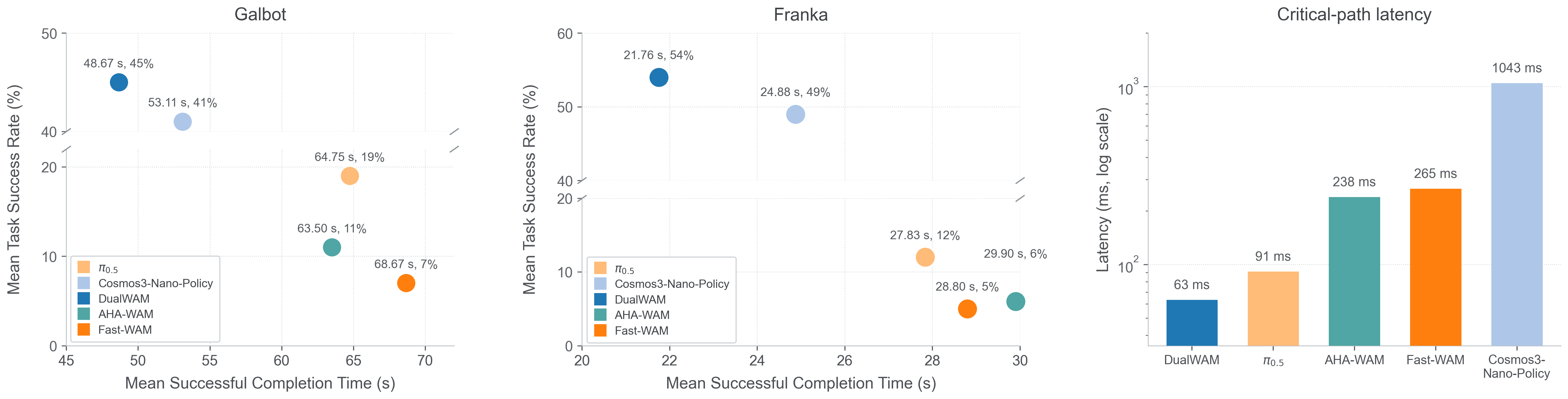}
\caption{\textbf{Zero-shot real-robot performance and critical-path latency.}
Left and center: full-task success rate versus mean completion time over successful rollouts on Galbot and Franka, respectively. Completion time is the number of executed control steps divided by the control frequency. Right: critical-path latency on a logarithmic scale. \method attains the highest success rate on both embodiments while providing the lowest latency and completion time.}
\label{fig:main-results}
\end{figure}

We first assess \method's zero-shot task performance and execution efficiency
across Franka and Galbot. Figure~\ref{fig:main-results} shows a consistent
advantage across both embodiments. \method achieves the highest success rates, reaching 54\% on Franka and 45\% on Galbot, outperforming Cosmos3-Nano-Policy by 5 and 4 percentage points, respectively.

Beyond task success, \method also completes successful episodes faster.
Mean physical completion time decreases from 24.88\,s to 21.76\,s on Franka
and from 53.11\,s to 48.67\,s on Galbot. These gains are consistent with
\method's combination of global planning and responsive local refinement.
Frequent local updates allow execution deviations to be corrected before they
accumulate into task failures or require substantial corrective motion,
supporting both higher success rates and shorter physical execution times.

\begin{figure}[t]
\centering
\begin{subfigure}{\linewidth}
  \centering
  \includegraphics[width=\linewidth]{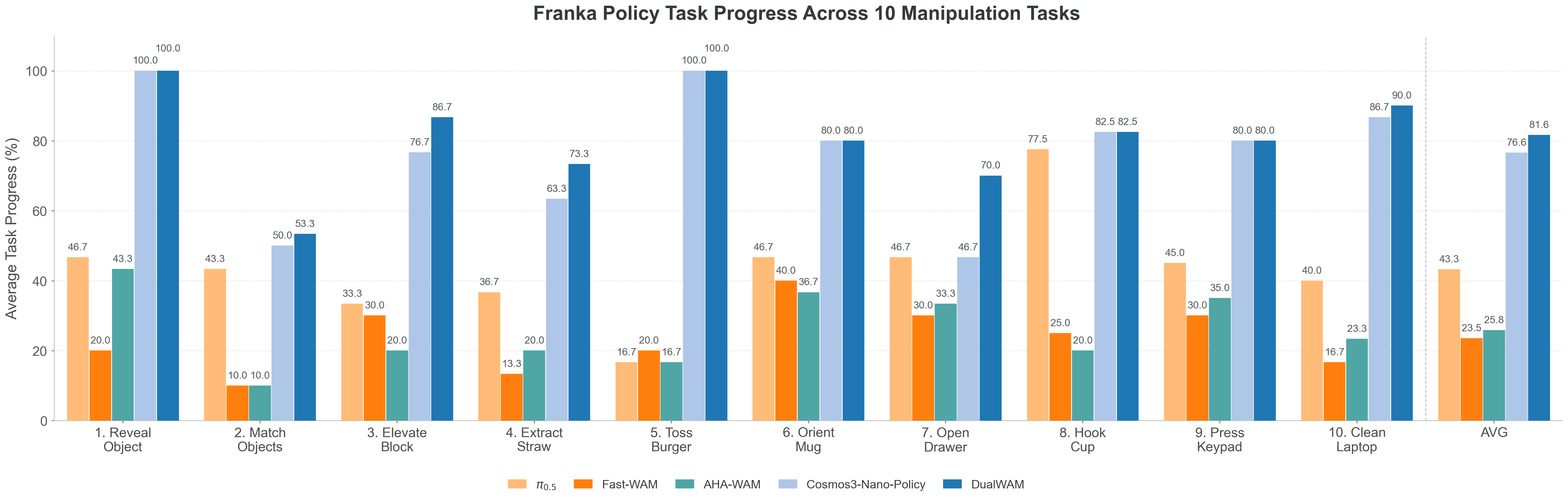}
  \phantomsubcaption\label{fig:franka-task-progress}
\end{subfigure}
\vspace{2pt}
\begin{subfigure}{\linewidth}
  \centering
  \includegraphics[width=\linewidth]{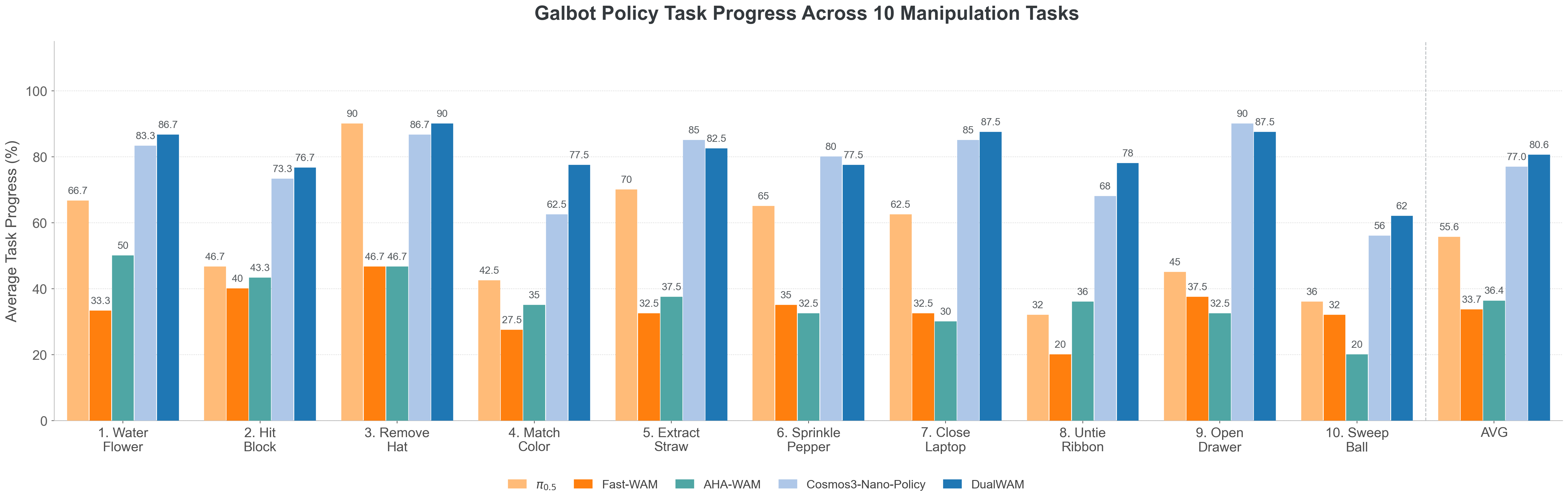}
  \phantomsubcaption\label{fig:galbot-task-progress}
\end{subfigure}
\caption{\textbf{Per-task zero-shot progress.} Mean normalized task progress on Franka (top) and Galbot (bottom). \method matches or exceeds the strongest baseline across all Franka tasks, while its Galbot gains are distributed across diverse interactions rather than being driven by a single task.}
\label{fig:per-task-progress}
\end{figure}

At the same time, \method achieves a
critical-path latency of only 63\,ms, the lowest among all evaluated policies.
Compared with 1043\,ms for Cosmos3-Nano-Policy, this corresponds to a
16.6$\times$ critical-path speedup. Together, these results
show that \method improves zero-shot task success while enabling more direct
closed-loop execution with substantially lower critical-path latency. Detailed per-task
results are provided in
Figures~\ref{fig:franka-task-progress} and~\ref{fig:galbot-task-progress}.

\FloatBarrier

\subsection{Ablation of dual-system decomposition and local world-action modeling}

We ablate whether global planning and local refinement should be separated and
whether the local refiner should retain explicit future prediction.
\emph{System 2 only} and \emph{System 1 only} use Cosmos3-Nano-Policy and
Cosmos3-Edge-Policy, respectively, as standalone all-view WAMs. The three
coupled variants use the same \sysone backbone, wrist observations, action
representation, and training data, differing only in how future visual
dynamics are modeled. \emph{Action-only DiT S1} removes future visual modeling
and predicts only actions; \emph{Fast-WAM S1} retains video--action co-training
but omits future-video prediction at inference~\citep{yuan2026fastwam}; and
\method jointly refines future wrist dynamics and actions from the temporally
aligned portion of \systwo's intermediate trajectory. This controlled
comparison separates the effect of training-time video modeling from that of
explicitly retaining future prediction during local refinement.

\begin{table}[H]
\centering
\caption{\textbf{Ablation of dual-system decomposition and local world-action
modeling on Franka.}
\method achieves the highest task progress and success rate at only 63\,ms
critical-path latency. Performance drops substantially when either the dual-system
decomposition or local world-action modeling is removed, highlighting the
importance of both global--local specialization and WAM-based local refinement.}
\label{tab:system-ablation}
\setlength{\tabcolsep}{5pt}
\renewcommand{\arraystretch}{1.08}
\begin{tabular}{lccc}
\toprule
Variant & Task Progress $\uparrow$ & SR $\uparrow$ & Critical-path Latency (ms) $\downarrow$ \\
\midrule
System 2 only & 76.6 & 49 & 1043 \\
System 1 only & 41.0 & 9 & 246 \\
System 2 + Action-only DiT S1 & 34.0 & 7 & 59 \\
System 2 + Fast-WAM S1 & 47.3 & 14 & 62 \\
\rowcolor{slowteal!8}
\method{} (ours) & 81.6 & 54 & 63 \\
\bottomrule
\end{tabular}
\end{table}

Compared with \emph{System 2 only}, \method raises task progress from 76.6 to
81.6 and success from 49\% to 54\%, while reducing latency from 1043 to
63\,ms, a 16.6$\times$ speedup. \emph{System 1 only} achieves 41.0 task progress
and 9\% success. Together, these results support combining low-frequency global
planning with high-frequency local refinement in two specialized systems.

With the global planner fixed, replacing \sysone with Action-only DiT or
Fast-WAM reduces success from 54\% to 7\% and 14\%, respectively, while
leaving local inference latency nearly unchanged (59\,ms and 62\,ms versus
63\,ms). The improvement from Action-only DiT to Fast-WAM suggests that
video--action co-training provides useful visual-dynamics priors even without
explicit future prediction at inference. However, the substantial remaining
gap to \method indicates that training-time video modeling alone does not
recover the benefit of retaining future prediction during local refinement.
By jointly predicting future wrist dynamics and actions, \method couples local
action refinement to its expected interaction outcomes, while the temporally
aligned intermediate state keeps these corrections consistent with the global
trajectory established by \systwo. This effect may be particularly important
in our zero-shot setting, where explicit future prediction can help leverage
pretrained visual-dynamics priors when task-specific action mappings are
unavailable.

\subsection{Asynchronous Execution and Edge--Cloud Deployment}

We next study two systems-level consequences of decoupling global planning
from local refinement: whether asynchronous execution improves closed-loop
control, and whether the same decomposition provides an efficient interface
for edge--cloud deployment. For the execution ablation, we keep the two WAMs,
model weights, and denoising schedule fixed, changing only how their updates
are coordinated. Synchronous execution requires a fresh \systwo intermediate
state for every \sysone action chunk, whereas asynchronous execution runs
\systwo in the background and lets \sysone refine from the latest valid state.

\begin{table}[H]
\centering
\caption{\textbf{Synchronous versus asynchronous execution on Galbot.}
With the same models and denoising schedule, synchronous execution requires a
fresh \systwo intermediate state for every \sysone action chunk, whereas
asynchronous execution reuses the latest valid state while \systwo runs in the
background. Asynchronous execution reduces critical-path latency from
850\,ms to 63\,ms (13.5$\times$) while improving task progress and success rate.}
\label{tab:async-execution}

\footnotesize
\setlength{\tabcolsep}{5pt}
\renewcommand{\arraystretch}{1.08}

\begin{tabular}{lccc}
\toprule
Execution & Avg. Progress $\uparrow$ & SR $\uparrow$ &
Critical-path Latency (ms) $\downarrow$ \\
\midrule
Synchronous & 73.3 & 39 & 850 \\
\rowcolor{slowteal!8}
Asynchronous (ours) & 80.6 & 45 & 63 \\
\bottomrule
\end{tabular}
\end{table}

\begin{table}[H]
\centering
\caption{\textbf{Edge--cloud communication on Galbot.}
Both methods synchronize at the same rate of 0.469\,Hz. \method reduces
the downlink payload per synchronization and average downlink bandwidth by
104.6$\times$, while reducing the 10~GbE payload-transfer bound by
18.0$\times$ compared with AHA-WAM.}
\label{tab:edge-cloud}

\scriptsize
\setlength{\tabcolsep}{2.5pt}
\renewcommand{\arraystretch}{1.08}

\resizebox{\linewidth}{!}{%
\begin{tabular}{lccccc}
\toprule
Method &
Sync Rate (Hz) &
Uplink / Sync $\downarrow$ &
Downlink / Sync $\downarrow$ &
Avg. Bandwidth (Up/Down) $\downarrow$ &
10GbE Transfer Bound $\downarrow$ \\
\midrule
AHA-WAM~\citep{cai2026ahawam}
& 0.469
& 0.737 MB
& 44.2 MB
& 0.346 / 20.7 MB/s
& 36.0 ms \\
\rowcolor{slowteal!8}
\method{}
& 0.469
& 2.07 MB
& 0.423 MB
& 0.972 / 0.198 MB/s
& 2.00 ms \\
\bottomrule
\end{tabular}}
\end{table}

Table~\ref{tab:async-execution} shows that asynchronous execution improves both
task performance and inference efficiency. Average task progress increases
from 73.3 to 80.6 and success rate from 39\% to 45\%, while critical-path
latency decreases from 850\,ms to 63\,ms, a 13.5$\times$ speedup. Under synchronous
execution, every \sysone action chunk must wait for a newly generated
\systwo intermediate state, placing both stages on the inference path.
Asynchronous execution removes this dependency: \systwo runs in the background,
while \sysone immediately refines actions from the latest available global
state using fresh wrist observations.

The higher success rate is consistent with more responsive local updates that
correct execution deviations using fresh feedback. Reusing each global state
across several local refinements may also help maintain a stable global intent
by reducing variation from repeated global-plan generation.

Table~\ref{tab:edge-cloud} further evaluates the communication cost of
splitting the global and local systems across cloud and edge. Both AHA-WAM
and \method synchronize at the same rate of 0.469\,Hz, isolating the effect
of the information exchanged between the two systems. AHA-WAM transfers
44.2\,MB of layer-wise video K/V context per synchronization, whereas
\method communicates only 0.423\,MB of coarse wrist latent and action prior,
reducing the downlink payload by 104.6$\times$. Accordingly, average downlink
bandwidth decreases from 20.7 to 0.198\,MB/s, also a 104.6$\times$
reduction. The larger uplink payload of \method mainly results from the
higher-resolution Galbot observation. Using the total bidirectional payload,
the corresponding 10~GbE transfer bound decreases from 36.0\,ms to
2.00\,ms, an 18.0$\times$ reduction. Since the two methods synchronize at
the same frequency, these gains directly reflect the substantially more
compact global--local communication interface of \method.

\subsection{Can DualWAM Benefit from Role-Matched Auxiliary Data?}

The specialized observation interfaces of \method provide natural entry points
for heterogeneous embodied data. We augment \systwo with egocentric videos
and \sysone with wrist-centric UMI demonstrations, first independently and
then jointly, on top of the same robot training data.

\begin{table}[H]
\centering
\caption{\textbf{Role-matched auxiliary-data ablation on Galbot.}
All variants are trained with robot trajectories. When auxiliary data are
used, egocentric videos are added only to \systwo and UMI demonstrations only
to \sysone, each mixed with robot data at a 1:1 ratio. Both sources improve
performance individually, and their combination achieves the best result.}
\label{tab:data-ablation}

\setlength{\tabcolsep}{5pt}
\renewcommand{\arraystretch}{1.08}

\begin{tabular}{llcc}
\toprule
\multicolumn{2}{c}{Training Data} &
\multicolumn{2}{c}{Performance} \\
\cmidrule(lr){1-2}\cmidrule(lr){3-4}
\systwo & \sysone & Task Progress $\uparrow$ & SR $\uparrow$ \\
\midrule
Robot & Robot & 80.6 & 45 \\
Robot + Ego & Robot & 86.0 & 54 \\
Robot & Robot + UMI & 84.8 & 51 \\
\rowcolor{slowteal!8}
Robot + Ego & Robot + UMI & \textbf{89.6} & \textbf{59} \\
\bottomrule
\end{tabular}
\end{table}

Table~\ref{tab:data-ablation} shows that both auxiliary data sources provide
consistent gains. Adding egocentric data to \systwo increases task progress
from 80.6 to 86.0 and success rate from 45\% to 54\%, while adding UMI data
to \sysone increases them to 84.8 and 51\%, respectively. Combining both
sources achieves the best performance, reaching 89.6 task progress and 59\%
success rate, an improvement of 9.0 progress points and 14 percentage points
in success over robot-only training. These complementary gains show that the
global--local specialization of \method also provides a natural interface for
incorporating heterogeneous, role-aligned embodied data without changing the
core architecture.

\section{Discussion and Limitations}

\method is designed for settings where manipulation benefits from global task organization and high-frequency local execution. It may be less useful when tasks are purely reactive, when a single camera already captures both global task context and local contact geometry, or when the high-noise system fails to establish a useful world-action state. The key distinction from closely related asynchronous or dual-system WAMs is the division of one denoising trajectory by semantic role, together with the wrist-only observation space used below the noise threshold.

Another limitation is compute. A two-system WAM can be more expensive to train than a compact policy-only baseline. The practical value of the design depends on whether \systwo can run sparsely enough that \sysone maintains the desired control rate. For this reason, latency and stale-plan robustness should be treated as primary results rather than implementation details.

\section{Conclusion}

We presented \method, an asynchronous dual-system WAM that assigns complementary intervals of one world-action denoising trajectory to global planning and local refinement. \systwo organizes a global high-noise state, while compact wrist-only \sysone continues low-noise local refinement with fresh interaction feedback. Across zero-shot tasks on Franka and Galbot, this separation improves both success and inference latency over existing policies. Targeted studies further show that the same decomposition can incorporate role-matched ego and UMI data and communicate efficiently in an edge--cloud setting. These are supporting benefits of a design whose central contribution is coordinating two WAMs around distinct planning and execution roles.

\bibliographystyle{assets/plainnat}
\bibliography{references}

\appendix
\section{Implementation Details}
\label{sec:additional-implementation-details}

Table~\ref{tab:implementation-details} summarizes the model, training, and deployment settings used in our experiments. Both systems are trained on complementary intervals of the same rectified-flow timestep axis. At deployment, the partially denoised trajectory produced by \systwo is divided into temporally aligned windows that \sysone continues denoising.

\begin{table}[!htbp]
\centering
\caption{\textbf{Implementation settings.}}
\label{tab:implementation-details}
\small
\setlength{\tabcolsep}{5pt}
\renewcommand{\arraystretch}{1.08}
\begin{tabularx}{0.92\linewidth}{lX}
\toprule
Setting & Value \\
\midrule
Global horizon $H_2$ & 32 frames \\
Local horizon $H_1$ & 8 frames \\
Training low-noise upper bound $\tau_{\mathrm{train}}$ & 0.7 \\
\systwo refresh interval & 32 control frames \\
Optimizer & AdamW \\
Learning rate & $2 \times 10^{-4}$ \\
Training precision & bfloat16 \\
Training iterations & 10k \\
Global batch size & 1,024 \\
Training hardware & 8 nodes $\times$ 8 NVIDIA H200 GPUs \\
Deployment hardware & 3 NVIDIA GeForce RTX 4090 GPUs \\
\bottomrule
\end{tabularx}
\end{table}

\paragraph{Logging.}
Every rollout stores planner-view frames, local wrist frames, the world-action state at the noise threshold, completed wrist-action windows, executed actions, global-state age, denoising steps, and per-module latency. These logs distinguish failures caused by poor global organization, stale global trajectories, and local execution errors.

\section{Zero-Shot Evaluation Details}

We evaluate ten unseen tasks on each embodiment and report Franka and Galbot separately, following DreamZero~\citep{ye2026dreamzero}. Each rollout receives a normalized task-progress score in $[0,1]$ from observable, task-specific milestones. Full completion is assigned $1.0$ and counted as a success; incomplete rollouts receive the partial credit specified below. For multi-object tasks, credit accumulates across completed subgoals and is capped at $1.0$. We average rollout scores to obtain task progress, whereas SR is the fraction of rollouts with a score of $1.0$. This distinction rewards meaningful partial execution without conflating it with full-task completion.

\subsection{Franka Tasks}

Figure~\ref{fig:franka-task-gallery} shows the initial configurations used for the ten Franka tasks. They cover object discovery, relational matching, tool use, articulated-object interaction, and spatial reorientation. Table~\ref{tab:franka-tasks} gives the exact language prompt and the observable milestones used for task-progress scoring.

\begin{table}[!htbp]
\centering
\caption{\textbf{Franka task prompts and progress criteria.} Partial scores are assigned only when full completion is not reached.}
\label{tab:franka-tasks}
\scriptsize
\setlength{\tabcolsep}{3.5pt}
\renewcommand{\arraystretch}{1.10}
\begin{tabularx}{\linewidth}{@{}p{0.15\linewidth}p{0.30\linewidth}X@{}}
\toprule
Task Name & Prompt & Task-Progress Criteria \\
\midrule
Reveal Object & Reveal the object under the cup. & $0.33$: touch the correct cup; $0.67$: move it; $1.0$: reveal the hidden object. \\
Match Objects & Match the objects with their corresponding boxes based on length. & $+0.16$: pick an object without placing it; $+0.16$: place each object in its correct box; $1.0$: match all objects. \\
Elevate Block & Elevate the block to the highest platform. & $0.33$: pick the block; $0.67$: place it on a platform; $1.0$: place it on the highest platform. \\
Extract Straw & Extract the straw from the cup. & $0.33$: touch the straw; $0.67$: move it; $1.0$: fully extract it from the cup. \\
Toss Burger & Toss the burger in the pan. & $0.33$: grasp the pan handle; $0.67$: move the burger; $1.0$: toss the burger in the pan. \\
Orient Mug & Orient the mug with its handle pointing toward the pot. & $0.33$: touch the mug; $0.67$: rotate it; $1.0$: orient the handle toward the pot. \\
Open Drawer & Open the top drawer. & $0.33$: touch the drawer; $0.67$: open it halfway; $1.0$: fully open it. \\
Hook Cup & Hook the cup onto the rack. & $0.25$: touch the cup; $0.50$: pick it up; $0.75$: approach the rack; $1.0$: hook the cup securely. \\
Press Keypad & Press the number pad and Enter on the keyboard. & $0.50$: press the keyboard; $1.0$: press the requested keys correctly. \\
Clean Laptop & Clean the laptop with a brush. & $0.33$: pick up the brush; $0.67$: bring it to the laptop; $1.0$: brush the laptop surface. \\
\bottomrule
\end{tabularx}
\end{table}

Figure~\ref{fig:franka-task-progress} disaggregates the aggregate result in the main paper. \method matches or exceeds the strongest baseline across all ten tasks, with particularly large gains on block elevation, straw extraction, and drawer opening, where maintaining a global objective while correcting local interactions is important.

\subsection{Galbot Tasks}

The Galbot suite in Figure~\ref{fig:galbot-task-gallery} emphasizes bimanual and contact-rich behaviors, including tool use, pouring, sweeping, untying, and articulated-object manipulation. The corresponding prompts and scoring milestones are listed in Table~\ref{tab:galbot-tasks}.

\begin{table}[!htbp]
\centering
\caption{\textbf{Galbot task prompts and progress criteria.} Milestones reflect increasing completion of the commanded interaction.}
\label{tab:galbot-tasks}
\scriptsize
\setlength{\tabcolsep}{3.5pt}
\renewcommand{\arraystretch}{1.10}
\begin{tabularx}{\linewidth}{@{}p{0.15\linewidth}p{0.30\linewidth}X@{}}
\toprule
Task Name & Prompt & Task-Progress Criteria \\
\midrule
Water Flower & Water the flower. & $0.33$: touch the kettle handle; $0.67$: lift the kettle; $1.0$: position it over the flower and perform a pouring motion. \\
Hit Block & Hit the block with the hammer. & $0.33$: touch the hammer; $0.67$: lift it; $1.0$: strike the block. \\
Remove Hat & Remove the hat from the hook. & $0.33$: touch the hat; $0.67$: grasp it; $1.0$: remove it from the hook. \\
Match Color & Put the toy pig into the corresponding plate based on its color. & $0.25$: touch the toy; $0.50$: pick it up; $0.75$: place it on an incorrect plate; $1.0$: place it on the color-matched plate. \\
Extract Straw & Extract the straw from the cup. & $0.25$: touch the straw; $0.50$: lift it; $0.75$: move it without successful extraction; $1.0$: fully extract it. \\
Sprinkle Pepper & Sprinkle pepper on the burger. & $0.25$: touch the shaker; $0.50$: grasp it; $0.75$: move it over the burger; $1.0$: tilt and sprinkle. \\
Close Laptop & Close the laptop. & $0.25$: touch the lid; $0.50$: initiate closing; $0.75$: partially close it; $1.0$: close it fully. \\
Untie Ribbon & Untie the ribbon on the gift. & $0.20$: touch with one hand; $0.40$: touch with both hands; $0.60$: grasp with one hand; $0.80$: grasp with both; $1.0$: pull the ribbon open. \\
Open Drawer & Open the lower drawer of the cooker. & $0.25$: touch the handle; $0.50$: grasp it; $0.75$: open halfway; $1.0$: fully open the drawer. \\
Sweep Ball & Sweep the paper balls into the dustpan. & $0.20$: touch the brush; $0.40$: lift it; $0.60/0.80/1.0$: sweep one/two/three balls into the dustpan. \\
\bottomrule
\end{tabularx}
\end{table}

Figure~\ref{fig:galbot-task-progress} shows that \method's improvement is distributed across diverse interaction types rather than being driven by a single task. The largest margins appear on color matching, ribbon untying, and sweeping, which benefit from repeated local correction over a longer task structure.

\end{document}